\documentclass{article}

\usepackage{microtype}
\usepackage{graphicx}
\usepackage{subfigure}
\usepackage{booktabs} % for professional tables

\usepackage{amsmath}
\usepackage{amssymb}
\usepackage{amsthm}

\newtheorem{theorem}{Theorem}

\newtheorem{lemma}[theorem]{Lemma}

\DeclareMathOperator*{\maximize}{maximize}

\usepackage{hyperref}

\usepackage[accepted]{icml2020}

\icmltitlerunning{Constrained Thompson Sampling for Wireless Link Optimization}

\begin{document}

\twocolumn[
\icmltitle{Constrained Thompson Sampling for Wireless Link Optimization}



\icmlsetsymbol{equal}{*}

\begin{icmlauthorlist}
\icmlauthor{Vidit Saxena}{kth,eric}
\icmlauthor{Joseph E. Gonzalez}{ucb}
\icmlauthor{Ion Stoica}{ucb}
\icmlauthor{Hugo Tullberg}{eric}
\icmlauthor{Joakim Jald\'{e}n}{kth}
\end{icmlauthorlist}

\icmlaffiliation{kth}{KTH Royal Institute of Technology, Stockholm, Sweden}
\icmlaffiliation{ucb}{University of California, Berkeley, USA}
\icmlaffiliation{eric}{Ericsson Research, Stockholm, Sweden}

\icmlcorrespondingauthor{Vidit Saxena}{vidits@kth.se}

\icmlkeywords{Machine Learning, ICML}

\vskip 0.3in
]



\printAffiliationsAndNotice{}  % leave blank if no need to mention equal contribution

\begin{abstract}

Wireless communication systems operate in complex time-varying environments. Therefore, selecting the optimal configuration parameters in these systems is a challenging problem. For wireless links, \emph{rate selection} is used to select the optimal data transmission rate that maximizes the link throughput subject to an application-defined latency constraint. We model rate selection as a stochastic multi-armed bandit (MAB) problem, where a finite set of transmission rates are modeled as independent bandit arms. For this setup, we propose Con-TS, a novel constrained version of the Thompson sampling algorithm, where the latency requirement is modeled by a high-probability linear constraint. We show that for Con-TS, the expected number of constraint violations over $T$ transmission intervals is upper bounded by $O(\sqrt{KT})$, where $K$ is the number of available rates. Further, the expected loss in cumulative throughput compared to the optimal rate selection scheme (i.e., the \emph{regret}) is also upper bounded by $O(\sqrt{KT \log K})$. Through numerical simulations, we demonstrate that Con-TS significantly outperforms state-of-the-art bandit schemes for rate selection.

\end{abstract}

\section{Introduction}
Transmitting information over a wireless channel is a complex phenomenon. The channel capacity is determined by the signal attenuation and baseband phase rotation, which in turn depend on a variety of parameters, including the signal's frequency, the distance between the transmitter and the receiver, and the nearby objects (\emph{scatterers}) that interact with the electromagnetic waves in the intervening period between transmission and reception. Furthermore, any of the transmitter, the receiver, or the scatterers might be mobile. As a result, the response of the wireless channel is time-varying and \emph{a priori} unknown~\cite{B_Wireless_Molisch}. Therefore to optimize a wireless link, the communication system needs to navigate its complex configuration parameter space, and do it quickly.

The central goal of wireless link optimization is to maximize the data throughput, since the wireless channel is a finite and expensive resource. At the same time, ubiquitous wireless services such as telephony and video streaming need to maintain a minimum service-level \emph{latency} to meet their respective quality of service (QoS) requirements. The goals of throughput maximization and latency are often in conflict: aggressive data transmission rates tend to improve throughput, but they also suffer from more frequent packet errors which hurts latency. However, state-of-the-art wireless link optimization schemes do not account for latency when selecting the optimal data transmission rates. As a result, the wireless link is often configured sub-optimally in terms of latency. In a recent field study of WiFi access points~\cite{pei2016wifi}, the WiFi hop was observed to cause over 60\% of the median round trip network latency. Further, higher WiFi hop latencies were found to be strongly correlated with packet retransmissions at the medium access control (MAC) layer. 

In this paper, we address throughput optimization for wireless links jointly with latency. The application-level latency requirement translates to a desired minimum MAC-level packet success rate to keep the packet retransmissions in check. We model this requirement through a linear constraint that must be satisfied with a high probability in every transmission interval. Subsequently, we formulate a \emph{rate selection} problem to learn the optimal data transmission rates over a sequence of transmission time intervals and propose a new algorithm, \emph{Con-TS}, based on the well-known Thompson Sampling (TS) heuristic. For our algorithm, we obtain finite-time bounds on the cumulative number of expected constraint violations as well as the cumulative loss in average throughput compared to the stationary optimal policy (i.e., the \emph{regret}). We numerically compare Con-TS with state-of-the-art bandit schemes proposed recently in the literature. Our results indicate that for latency-constrained rate selection, Con-TS significantly outperforms the existing approaches, and is attractive for online optimization problems that deal with probabilistic linear constraints.

\textbf{Our Contributions}: 

(i) We formulate rate selection as a linearly constrained optimization problem and show that it is equivalent to the bandits with knapsack problem studied in~\cite{badanidiyuru2013bandits}. However, in contrast to the knapsack problem that deals with strict budgets, latency is a \emph{soft} requirement in the sense that it needs to be satisfied only in expectation. Consequently, we apply a linear programming (LP) relaxation that can be solved in polynomial time using commonly available numerical solvers. %Similar soft constraints have recently been proposed in~\cite{meisami2018sequential} within an online advertising setting.

(ii) For the constrained bandit problem, we propose Con-TS, a probabilistically constrained version of the TS heuristic originally introduced in~\cite{thompson1933likelihood}. We obtain the first prior-independent, finite-time performance bounds for Con-TS. These bounds scale as $O(\sqrt{KT})$ for the cumulative expected constraint violations as well as the cumulative loss in throughput compared to the best rate selection policy, where $K$ and $T$ denote the number of available rates and the transmission time intervals respectively.

(iv) Using a range of established synthetic WiFi scenarios, we compare Con-TS with two recently proposed bandit schemes: unimodal Thompson Sampling (UTS) proposed in~\cite{paladino2017unimodal} and a constrained upper confidence bound scheme, Con-KL-UCB, which uses Kullback-Liebler bounds and is a simple extension of the scheme proposed in~\cite{chen2018beyond} for online weblink selection. Through numerical experiments, we demonstrate that Con-TS significantly outperforms both the previous schemes in terms of the throughput-violation ratio and the cumulative number of constraint violations.

\section{Related Work}
TS was first introduced in~\cite{thompson1933likelihood} and is a well-known Bayesian heuristic for solving sequential learning problems. In contrast to the frequentist upper confidence bound (UCB) schemes, TS assigns prior distributions over the MAB parameters, which are refined sequentially over the course of the experiment. In ~\cite{chapelle2011empirical}, the authors demonstrated attractive empirical results using TS for several common bandit problems, sparking renewed interest in the technique. The empirical results were augmented by the extensive theoretical work in~\cite{kaufmann2012thompson} and ~\cite{agrawal2013further} that established asymptotically optimal and prior-dependent $O(\log T)$ bounds on the TS regret. For MAB settings with hard linear constraints, an extension to TS was recently proposed in~\cite{ferreira2018online} for which prior-dependent regret bounds that scale as $O(\sqrt{T})$ were developed.  In contrast to~\cite{ferreira2018online} that dealt with hard linear constraints and a single performance metric (i.e., the regret), our problem setting has a soft (i.e., probabilistic) linear constraint and consequently involves two performance metrics: throughput and constraint violation. Hence, we develop new bounds for each of these metrics, which scale as $O(\sqrt{T})$.

Rate selection is closely related to important problems in otherwise unrelated domains, such as online advertising and dynamic pricing. The online ad publishing problem was studied in~\cite{chen2018beyond},where a publishing entity seeks to maximize the cumulative ad revenue generated from user clicks. At the same time, the publisher wants to ensure that unattractive, high-revenue ads do not end up cluttering the webpage displayed to a user. The publisher balances these conflicting goals by estimating the click-through-rate (CTR) for each ad and down-selecting the ads that have an insufficient CTR~\cite{mookerjee2016optimizing}. In~\cite{chen2018beyond}, a probabilistically constrained, multi-play UCB approach, ConUCB, was developed for determining the optimal subset of links that are displayed to the user in each round. The ConUCB algorithm learns the \emph{first-level} click-through-rate (CTR) for each link, and the~\emph{second-level} expected revenue collected after a user clicks a particular link. Subsequently, ConUCB calculates a probabilistic selection vector in each round to sample an optimal subset of links, i.e., the subset of links which exceed the CTR constraint and maximize the expected cumulative revenue over a finite time horizon. For ConUCB, regret and violation bounds that scale as $O(\sqrt{T})$ were also developed. The rate selection problem studied in this paper can be interpreted as a single-play variant of the problem studied in~\cite{chen2018beyond}, where the second-level reward is the fixed throughput associated with a given rate. The bounds developed for Con-TS in this paper match the previously discovered bounds for ConUCB. For numerical comparison with Con-TS, we extend ConUCB with KL-divergence bounds which has been shown to outperform index-based UCB in~\cite{garivier2011kl}. Further, through numerical experiments, we show that Con-TS outperforms this improved algorithm, termed Con-KL-UCB, for the studied WiFi scenarios.

MAB based \emph{unconstrained} rate selection, where the goal is to learn the throughput-maximizing rate, is an area of active research for WiFi~\cite{combes2014optimal} as well as cellular networks~\cite{saxena2018deep, saxena2019contextual}. In~\cite{combes2014optimal}, an upper confidence bound (UCB) based, graphical optimal rate sampling (G-ORS) algorithm was proposed that exploits the fact that the throughput function over the rates has a unique maxima (unimodality property). A TS approach that also exploits the unimodality property, unimodal Thompson sampling (UTS), was subsequently proposed in~\cite{paladino2017unimodal} and was shown to outperform G-ORS type of algorithms. However, the existing bandit approaches do not consider latency requirements which is critical for practical wireless optimization. In this paper, we show that Con-TS significantly outperforms UTS in terms of the latency-constrained rate selection objective.

\section{Model and Objectives}
\label{sec:systemmodel}
\subsection{Wireless Link Model}

A wireless transmitter sends packetized data over discrete transmission intervals. In each transmission interval, the transmitter processes a variable number of data bits by first encoding them for reliable reception and subsequently modulating them onto waveforms for transmission over the air. In each transmission interval, the choice of the rate index, $k\in\{1,\dots,K\}$, determines the number of data bits $D_{k}$ that are prepared for transmission. The sequence of $D_{k}$ data bits are mapped to a sequence of $C_{k}$ coded bits by an error protecting channel code.  The coded bits are then mapped onto waveforms, represented as data symbols from a symbol alphabet, and transmitted over the air. Assuming that $\Delta f$ and $\Delta t$ is the bandwidth and transmission duration, respectively, the (instantaneous) rate of the transmission is given by
\begin{align}
r_{k} \triangleq \frac{1}{\Delta t \Delta f } D_{k} \quad \text{[bits / s / Hz]} \, ,\label{eq:rate}
\end{align}
i.e., the number of information bits transmitted in the transmission interval normalized by the resources utilized to do so.  The specific channel code used (type and rate), and the symbol alphabet (e.g., QPSK, 8-PSK or 16-QAM) is determined by the modulation and coding scheme $m_{k}$, thus allowing for adaptive control of the data rate and error protection of the transmission. Known, deterministic, pilot symbols are periodically inserted into the transmitted signal in order to allow the receiver to estimate the frequency- and time-dependent state of the wireless channel.

The wireless channel adds stochastic effects such as channel attenuation, baseband phase rotation, noise, and interference to the transmitted signal as it propagates over the air. The receiver attempts to recover the transmitted bits by first compensating for the estimated channel attenuation and phase rotation and subsequently applying inverse operations for the encoding and modulation steps performed prior to transmission. The receiver reports back to the transmitter a positive acknowledgement, $x_{k}=1$, in the event the reconstructed data bits are received correctly (usually validated by a cyclic redundancy check embedded in the packet), and a negative acknowledgement, $x_{k}=0$, otherwise. Viewing $x_k = x(r_k)$ as a family of stochastic variables, indexed by the rate $r_k$ and with distributions induced by the random channel effects, yields the packet success rate, $\mu_k=Pr\big\{ x(r_k)=1 \big\}$ and expected throughput, $r_k \cdot\mathbb{E}\big\{ x(r_k)=1 \big\}=r_k\mu_k$. The packet success rate is an \emph{a priori} unknown function of the transmission parameters and the wireless channel state. Consequently for optimal rate selection, the values $\mu_1,\dots,\mu_K$, must be learnt by sequentially probing the wireless channel with the available rates.

\subsection{Optimal Rate Selection}
\label{sec:opt_selection}

Consider a wireless link where $K$ discrete data transmission rates $r_1,\dots,r_K$ are available for packetizing the information bits. For the $k^\text{th}$ rate, the packet success probability is denoted by $\mu_k$, which is assumed to be unknown at the start of the experiment and fixed for the duration of the experiment. In each transmission interval, one of the available rates is selected for packetizing the information bits that are subsequently transmitted over the air. Naturally, a higher data transmission rate corresponds to a lower packet success probability since fewer error protection bits are included in the data packet. For the choice of rate index $k(t)\in\{1,\dots,K\}$ at time interval $t$, either $r_{k(t)}$ bits are delivered to the receiver with probability $\mu_{k(t)}$, or zero bits are delivered with probability $1-\mu_{k(t)}$. Therefore, the binary packet success event, $x_{k(t)}$, is an independent and identically distributed (i.i.d.) Bernoulli random variable with mean $\mu_{k(t)}$, i.e.,  $x_{k(t)}\sim~\text{Bern}(\mu_{k(t)})$.

Consider the sequence of transmission time intervals $t=1,\dots,T$. Our goal is to find the optimal sequence of rate indices $k(1),\dots,k(T)$ that maximizes the average link throughput subject to a configurable, fixed packet success constraint $\tau$,
\begin{align}
    &\maximize_{k(1),\dots,k(T)}\,\,\frac{1}{ T} \sum_{t=1}^{T}r_{k(t)}x_{k(t)} \nonumber \\
    &\quad\text{subject to}\,\,\frac{1}{ T}\sum_{t=1}^Tx_{k(t)}\geq \tau,
    \label{eq:opt_rate_sel}
\end{align}
where $x_{k(t)}\in \{0,1\},\,\forall\,t\in{1,\dots,T}$. The problem in~\eqref{eq:opt_rate_sel} is an instance of the NP-hard knapsack problem that has been been treated recently in the context of MAB problems~\cite{badanidiyuru2013bandits, agrawal2014bandits}. This prior work deals with hard constraints, for example when applied to revenue management where the constraint denotes a strict limit on the number of items available in the inventory. In contrast, the latency requirement for networked communication is a soft constraint. Occasional violations of the latency constraint are acceptable (and often unavoidable on account of dynamic nature of the wireless channel), as long as the average latency across multiple data flows stays below a configurable threshold. 

We relax the problem in \eqref{eq:opt_rate_sel} by applying a linear programming (LP) relaxation that assigns probability weights to each rate. This LP can solved in polynomial time to obtain a \emph{probabilistic rate selection vector}. The optimal rate selection strategy is therefore a probabilistic mixture of rates: with some probability, high data rates is selected that may violate the packet success constraint. Therefore, rest of the probability mass is assigned to a relatively low data rates so that, on average, the packet success constraint is satisfied. We can write the LP relaxation of~\eqref{eq:opt_rate_sel} as
\begin{align}
    p^*_{1,\dots,K}=&\maximize_{y_{1,\dots,K}}\,\,\sum_{k=1}^Ky_kr_k\mu_k \nonumber \\
    &\quad\text{subject to}\,\sum_{k=1}^Ky_k\mu_k\geq \tau, \nonumber \\
    &\qquad\qquad\quad\sum_{k=1}^Ky_k = 1,\nonumber \\
    &\qquad\qquad\quad y_k \geq 0.
    \label{eq:opt_rate_sel_linprog}
\end{align}
The probability weights assigned to each arm by the optimal solution to \eqref{eq:opt_rate_sel_linprog} are denoted by $p_1^*,\dots,p_K^*$ respectively. The vector of these probability weights constitutes the \emph{stationary optimal policy} for this problem. Subsequently, we formulate the LP in \eqref{eq:opt_rate_sel_linprog} as a MAB problem, where the arms of the bandit model the data transmission rates, and propose a TS-based online learning algorithm.

The true packet success probabilities, $\mu_1,\dots,\mu_K$, are generated by the environment and are \emph{a priori} unknown. At time interval $t$, the transmitter selects a single rate index $k(t)$ and observes the packet success or failure event $x_{k(t)}$. Therefore, the transmitter needs to trade-off between \emph{exploring} the available rates and \emph{exploiting} the packet success probabilities learnt so far. The goal of any MAB algorithm is to learn a policy that optimizes the rate selection problem in~\eqref{eq:opt_rate_sel_linprog}. We define the following measures of the constraint violation and the regret respectively, 
\begin{align}
    \mathcal{V}(T) &= \mathbb{E}\bigg[\bigg[T\tau - \sum_{t=1}^T\sum_{k=1}^Kp_k(t)\mu_k\bigg]_+\bigg], \nonumber \\
    \mathcal{R}(T) &= \mathbb{E}\bigg[\bigg[ T\sum_{k=1}^K p_k^*r_k\mu_k - \sum_{t=1}^T \sum_{k=1}^Kp_k(t)r_k\mu_k\bigg]_+\bigg],
    \label{eq:perf_metrics}
\end{align}  
where $[x]_+=\max(0,x)$ and $p_k(t)(t)$ is the probability mass assigned to rate index $k$ at time step $t$. The violation metric measures the expected number of constraint violations accumulated until time $T$. In contrast, the regret metric measures the expected loss in throughout compared to the stationary optimal policy. The two metrics are related in the sense that they must be optimized simultaneously by a MAB policy. However, the classical notion of regret is insufficient in the constrained problem addressed here: unconstrained MAB policies (such as UTS), which optimize for maximum throughput while ignoring the constraint, can achieve a trivially zero regret by always selecting a high data rate. Therefore, in order to study the performance of MAB algorithms in a constrained setting, we propose \emph{throughput-violation ratio} as a suitable evaluation metric, defined as
\begin{align}
      \mathcal{W}(T) &= \frac{\mathbb{E}\bigg[\sum_{t=1}^T \sum_{k=1}^Kp_k(t)r_k\mu_k\bigg]}{ \mathbb{E}\bigg[\bigg[T\tau - \sum_{t=1}^T\sum_{k=1}^Kp_k(t)\mu_k\bigg]_+\bigg]}.
\end{align}
A higher value of $\mathcal{W}(T)$ indicates better exploitation of the infrequent violations of the probabilistic constraint and vice versa. In general, $\mathcal{W}(T)$ encodes the reward-violation metric for probabilistically constrained MABs, which has also been proposed for the CTR-constrained weblink selection scenario in~\cite{chen2018beyond}.

\section{Constrained Thompson Sampling}
\label{sec:alg}
\subsection{Algorithm}

We first recall the classic TS algorithm for stochastic MABs and where the rewards across the MAB arms are distributed according to independent Bernoulli distributions with \emph{a priori} unknown means. TS starts by assigning independent, uniform, prior Beta distributions for each arm mean. At the beginning of each round, classical TS generates a sample from the Beta distributions for each MAB arm, plays the arm that returns the largest sample, and collects the reward generated by the environment. Subsequently, Bayes' rule is invoked to update the distribution for the played arm mean with the reward collected and the experiment proceeds to the next round. Although the choice of TS priors is arbitrary, using a Beta distribution follows naturally from a special relationship between the Beta and Bernoulli distributions: the product of a Beta-distributed and a Bernoulli-distributed random variable is also distributed according to a Beta distribution with new parameters. Therefore, by choosing a Beta prior for the arm mean, the calculation of the posterior distribution from the observed (Bernoulli-distributed) reward follows a simple update rule.

\begin{algorithm}[tb]
   \caption{Con-TS}
   \label{alg:con_ts}
\begin{algorithmic}[1]
   \STATE {\bfseries Input:} Rates $\{r_1,\dots,r_K\}$, Constraint $\tau$
   \STATE \textbf{Initialize:} $\alpha_{\{1,\dots,K\}}(0)=1, \beta_{1\dots K}(0)=1$.
   \FOR{ Transmission time interval $t=1$ {\bfseries to} $T$ }
   \STATE Sample $\widetilde{\mu}_k(t)\sim Beta(\alpha_k(t-1), \beta_k(t-1))$ for each arm $k=1,\dots,K$.
   \STATE Solve the linear program:
   \begin{align}
    p_{1,\dots,K}(t)=\maximize_{y_{1,\dots,K}}\,&\sum_{k=1}^K \bigg(r_k\widetilde{\mu}_k(t)\bigg)y_k \nonumber \\
     \quad\text{subject to} &\sum_{k=1}^K \widetilde{\mu}_k(t)y_k \geq \tau,\nonumber \\
     &\sum_{k=1}^Ky_k = 1,\text{ and }\nonumber \\
    &y_k \geq 0\quad \forall\,k\in\{1,\dots,K\}.
    \label{eq:Con-TS-alg}
    \end{align}
   \IF{\eqref{eq:Con-TS-alg} has a feasible solution,} 
    \STATE Sample $k(t)~\sim [p_1(t),\dots,p_K(t)]$
   \ELSE  % No feasible solution
    \STATE Sample $k(t)$ uniformly from $\{1,\dots,K\}$.
   \ENDIF
   \STATE \textbf{Observe:} Packet success $x_{k(t)}$.
   \STATE \textbf{Update:}
   \begin{align}
             &\alpha_k(t) = \alpha_{k(t)}(t) + x_{k(t)} \nonumber \\
             & \beta_k(t)=\beta_{k(t)}(t) + (1-x_{k(t)}).
             \label{eq:update_rule}
   \end{align}
   \ENDFOR
\label{alg:ts_linprog}
\end{algorithmic}
\end{algorithm}

For the rate selection problem addressed in this paper, the goal is to maximize the cumulative reward (i.e., the throughput) \emph{subject to} a linear constraint on the expected packet success rate. Hence, it is not sufficient to simply select the rate that maximizes the expected throughput in each transmission time interval. We propose the Con-TS algorithm that incorporates this linear constraint within the TS heuristic. At the beginning of the experiment, Con-TS assigns uniform Beta priors $Beta(\alpha_k(0)=1,\beta_k(0)=1),\,k=1,\dots,K$ to the each of the available data transmission rates, where $\alpha_k$ and $\beta_k$ are the parameters of the Beta distribution. The Beta distributions model the packet success probabilities that have the true, \emph{a priori} unknown values, $\mu_1,\dots,\mu_K$ respectively. In each transmission time interval $t=1,\dots,T$, Con-TS samples the packet success probability for each rate, $\widetilde{\mu}_k(t)\sim Beta(\alpha_k(t-1), \beta_k(t-1))$, which is used to parameterize the LP in~\eqref{eq:opt_rate_sel_linprog}. The solution to this LP is the probabilistic rate selection vector $p_1(t),\dots,p_K(t)$. Con-TS uses the rate index $k(t)$ obtained by sampling the probabilistic rate selection vector to packetize the information bits for transmission over the air. The packet success event is captured in the binary random variable $x_{k(t)}$, which has the (unknown) mean value $\mu_{k(t)}$. The posterior parameters of the Beta distribution $Beta(\alpha_{k(t)}(t), \beta_{k(t)}(t))$ are updated with the observed $x_{k(t)}$ using the update rule in~\eqref{eq:update_rule} and the experiment proceeds to the next round. The complete Con-TS algorithm is described in Alg.~\ref{alg:ts_linprog}.

\subsection{Non-stationary Environments}
\label{sec:nonstat}

The discussion so far has assumed stationary packet success probabilities. However, wireless channels are dynamic on account of moving terminals and surrounding objects. For the most common indoor wireless scenarios, the average data transmission only session lasts a few hundred milliseconds, far below the timescale for any significant physical movement. The packet success probabilities can then be considered approximately stationary. However, particularly in outdoor high-speed wireless environments, the wireless channel may change sufficiently fast to alter the packet success probabilities before the rate selection scheme has a chance to converge. To deal with non-stationary environments, we propose a sliding window protocol where only the historical data from a time window of configurable size is used for learning~\cite{garivier2008upper, combes2015dynamic}. Owing to space constraints, we provide the numerical results for non-stationary environments in the supplementary material.

\subsection{Structural Properties}

The rate selection problem exhibits a rich structure owing to the specific construction of wireless communication systems. In this paper, we do not exploit these structure properties to allow wider applicability of the proposed algorithm. However, we believe that possible future extensions to Con-TS may make exploit them for practical gains in the wireless communication scenario.

\textbf{Monotonicity: }
The packet success probability is a monotonic function of the data transmission rates. Selecting a higher transmission rate naturally leads to a lower mean packet success rate and vice versa. Intuitively, this relationship between the rates could be exploited for speeding up the the MAB learning. However, observing a packet success (failure) at time $t$ does not guarantee that a lower (higher) rate would have succeeded (failed) in that time interval, i.e., monotonicity only holds \emph{on average} and not instantaneously. Recently, an extension to TS was proposed in~\cite{gupta2019link} to exploit this property, where the sampling step was constrained to always return samples that are a monotonic function of the data transmission rates. However, the scheme in~\cite{gupta2019link} scales poorly in terms of the computational complexity and does not incorporate the probabilistic latency constraints addressed here.

\textbf{Unimodality: }
The expected throughput, $\mu_k\cdot r_k$, is known to be a unimodal function of the available data transmission rates. This property arises from the construction of the error-protecting channel codes and has previously been exploited in~\cite{combes2014unimodal}, which was generalized and extended to TS in~\cite{paladino2017unimodal}. In the numerical section, we show that Con-TS significantly outperforms the (unconstrained) UTS algorithm in most scenarios. However, for the case where a data transmission rate both  maximizes the throughput and satisfies the constraint, the constraint is satisfied trivially. In this setting, UTS has some performance benefit compared to Con-TS.

\subsection{Computational Complexity}

A typical wireless link optimization scheme needs to converge quickly, i.e. within a few milliseconds, to be of practical use. Con-TS introduces a LP subroutine in every TS step, which increases the computational complexity compared to TS. However, computationally efficient LP solvers, such as those based on Dantzig’s simplex algorithm, have long been used for real-time control \cite{murty1983linear}. Moreover, the size of the rate selection LP is relatively small (there are typically only a few tens of available data transmission rates). On commercial grade access points which are typically equipped with multi-core ARM processors, this LP can be solved in real time using the spare processing cycles.

\section{Theoretical Analysis}
\subsection{Preliminaries}

We develop Bayesian, prior-independent, upper bounds on the finite-time violation and regret metrics defined in~\eqref{eq:perf_metrics}. Recall that $\mu_1,\dots,\mu_K$ are the true packet success probabilities, $p_1^*,\dots,p_K^*$  denotes the stationary optimal solution to \eqref{eq:opt_rate_sel_linprog}, $\widetilde{\mu}_1(t),\dots,\widetilde{\mu}_K(t)$ are the sampled packet success probabilities at time step $t$, and $p_1(t),\dots,p_K(t)$ are the corresponding rate selection probabilities obtained by solving the LP in Alg.~\ref{alg:ts_linprog}. Let $\mathcal{H}(t)=\{k(1),x_k(1),\dots,k(t-1),x_k(t-1)\}$ denote the history of the rates selected and the observed packet success events until time $t$. In the Bayesian setting where the parameters of the problem are chosen by nature, the distribution of the sampled parameters conditioned on $\mathcal{H}(t)$ is same as the conditional distribution for the true probabilities, i.e., $P(\widetilde{\mu}_k(t)|\mathcal{H}(t))=P(\mu_k|\mathcal{H}(t)),\,k=1,\dots,K$~\cite{russo2014learning,ferreira2018online}. Subsequently, for any uniformly consistent LP solver, the rate selection probabilities conditioned on $\mathcal{H}(t)$ are also distributed similarly, \begin{align}
    P(p_k(t)|\mathcal{H}(t))=P(p_k^*|\mathcal{H}(t)),\,k=1,\dots,K.
\end{align}
Next, we define the following upper and lower confidence bounds introduced in~\cite{russo2014learning}:
\begin{align}
    U_k(t) &= \min{\bigg(1, \hat{\mu}_k(t)+\sqrt{\frac{\log_+{\frac{TK}{N_k(t-1)}}}{N_k(t-1)}}\bigg)} \\
    L_k(t) &= \max{\bigg(0, \hat{\mu}_k(t)-\sqrt{\frac{\log_+{\frac{TK}{N_k(t-1)}}}{N_k(t-1)}}\bigg)},
    \label{eq:conf_intv}
\end{align}
respectively, where $\log_{+}(x)=\log(x)\mathbf{1}_{\{x\geq 1\}}$, $N_k(t-1)$ is the number of times that rate $k$ has been selected at the beginning of time step $t$, and $\hat{\mu}_k(t) = \mathbb{E}[\mathbf{1}_{k(t) = k}]$ is the empirical mean success probability for rate $k$ at time $t$. From \cite{ferreira2018online} we we extract the following relationships between the true packet success probabilities and the confidence bounds:
\begin{lemma}
\cite{ferreira2018online}[Lemma EC.1] For any $k\in\{1,\dots,K\}$ and $t\in\{1,\dots,T\}$, the following inequalities hold:
\begin{align}
    &\mathbb{E}\big[(\mu_k-U_k(t))_+\big] \leq 6\frac{1}{\sqrt{KT}}, \text{ and}\\
    &\mathbb{E}\big[(L_k(t) - \mu_k)_+\big] \leq 6\frac{1}{\sqrt{KT}}.
\end{align}
\label{lemma_1}
\end{lemma}

\begin{lemma}
\cite{ferreira2018online}[Lemma EC.2] For any $k\in\{1,\dots,K\}$ and $t\in\{1,\dots,T\}$, the following inequalities hold:
\begin{align}
    &\sum_{t=1}^T\sum_{k=1}^K\mathbb{E}\big[(U_k(t)-\widetilde{\mu}_k(t))p_k(t)\big] \leq 12\sqrt{KT\log K}, \text{ and} \\
    &\sum_{t=1}^T\sum_{k=1}^K\mathbb{E}\big[(\widetilde{\mu}_k(t)-L_k(t))p_k(t)\big] \leq 12\sqrt{KT\log K}.
\end{align}
\label{lemma_2}
\end{lemma}

\subsection{Finite-Time Bounds}

\begin{theorem}
    The expected violation for Con-TS is upper bounded by
    \begin{align}
        \mathbb{E}\big[\mathcal{V}(T)\big] \leq 12\sqrt{KT} + O(K^2\log T\sqrt{T}).
    \end{align}
    \label{th:violation}
\end{theorem}

\emph{Sketch of the proof:} To bound the Bayesian violation probability, we estimate the expected number of times that the sampled packet success probability $\widetilde{\mu}_k(t)$ falls outside the upper and lower confidence intervals defined in~\eqref{eq:conf_intv} until time $T$. The central challenge arises from the stochastic nature of the sampling step in each round. To overcome this issue, we note that conditioned on the history $\mathcal{H}(t)$, the sampled and true packet success probabilities are distributed identically, i.e., $P(\widetilde{\mu}_k(t)|\mathcal{H}(t))=P(\mu_k|\mathcal{H}(t))$. We exploit this fact along with the bounds in Lemma~\ref{lemma_1} to control part of the violation bounds. For rest of the bound to work, we need a relationship between the expected violations accrued between the upper and lower bound, which was provided in~\cite{russo2014learning}. Combining the two, we obtain the result. Details of the proof are provided in Sec.~\ref{app:proof_vio}.

\begin{theorem}
    The expected regret for Con-TS is upper bounded by
    \begin{align}
        \mathbb{E}\big[\mathcal{R}(T)\big] \leq r_{\max}(6\sqrt{KT} + 12\sqrt{KT\log K}),
    \end{align}
where $r_{\max}=\max(\{r_1,\dots,r_K\})$.
\end{theorem}

\emph{Sketch of the proof:} This proof proceeds in a similar manner to the proof of Theorem~\ref{th:violation}. To handle the additional $r_k$ term, we note that the maximum regret accrued in any round is smaller than or equal to the maximum available data transmission rate rate, $r_{\max}$. Details of the proof are provided in Sec.~\ref{app:proof_regret}.

%Similarly, we have that $E\bigg[\sum_{t=1}^Tx_{k(t)}\bigg] = E\bigg[\sum_{t=1}^T\mu_kp_k(t)\bigg]$

\section{Numerical Results}
\label{sec:numrel}
\begin{figure*}[t!]
\centering

    \subfigure[Tput / Vio., Gradual]{
    \includegraphics[width=0.23\textwidth]{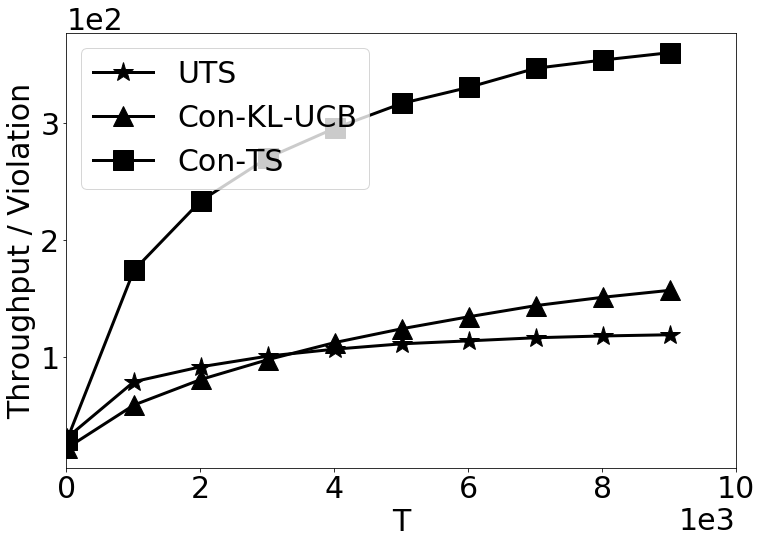}
        \label{fig:rew_gradual}
    }
    \subfigure[Tput / Vio., Lossy]{
    \includegraphics[width=0.23\textwidth]{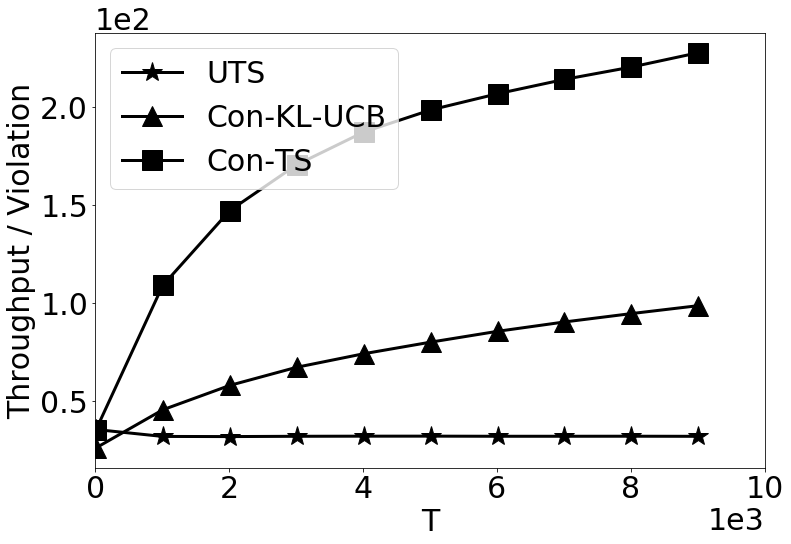}
        \label{fig:rew_lossy}
    }
    \subfigure[Tput / Vio., Steep ]{
    \includegraphics[width=0.23\textwidth]{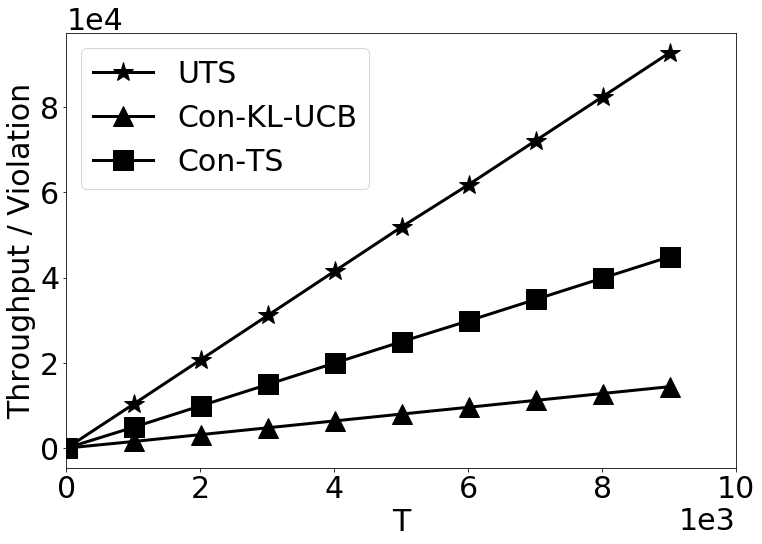}
        \label{fig:rew_steep}
    }
    \subfigure[Tput / Vio., Linear ]{
    \includegraphics[width=0.23\textwidth]{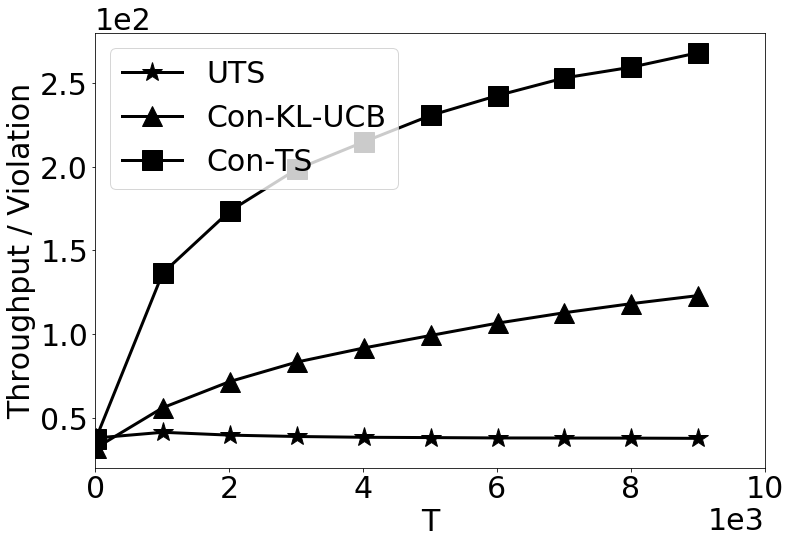}
        \label{fig:rew_linear}
    }
    
    \subfigure[Vio., Gradual]{
    \includegraphics[width=0.23\textwidth]{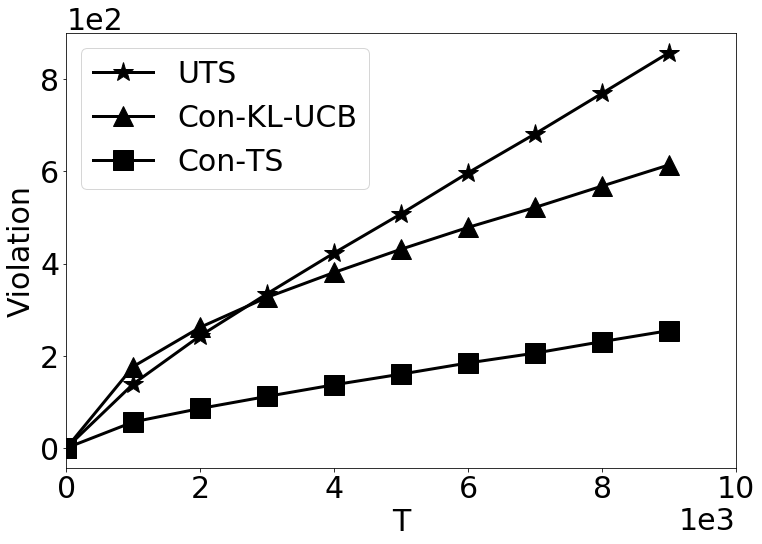}
        \label{fig:vio_gradual}
    }
    \subfigure[Vio., Lossy]{
    \includegraphics[width=0.23\textwidth]{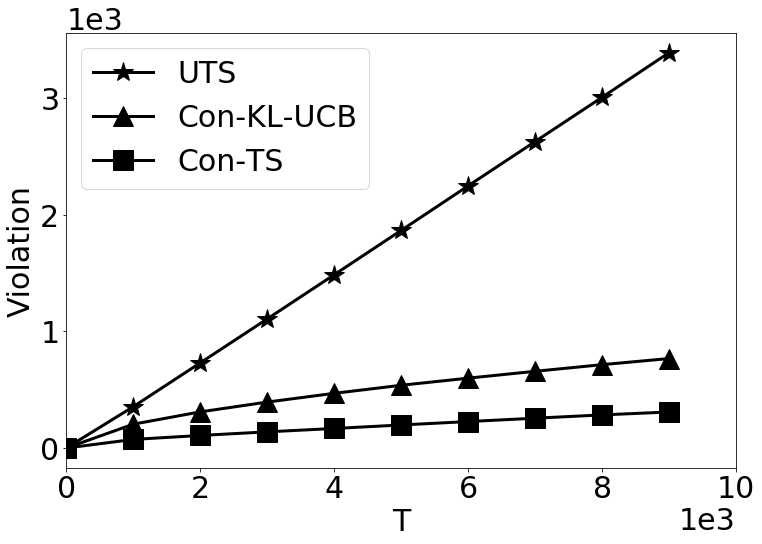}
        \label{fig:vio_lossy}
    }
    \subfigure[Vio., Steep]{
    \includegraphics[width=0.23\textwidth]{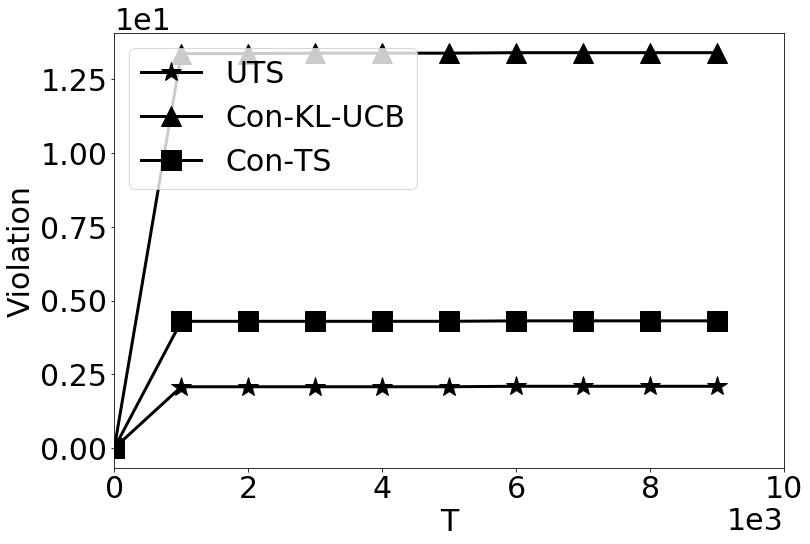}
        \label{fig:vio_steep}
    }
    \subfigure[Vio., Linear]{
    \includegraphics[width=0.23\textwidth]{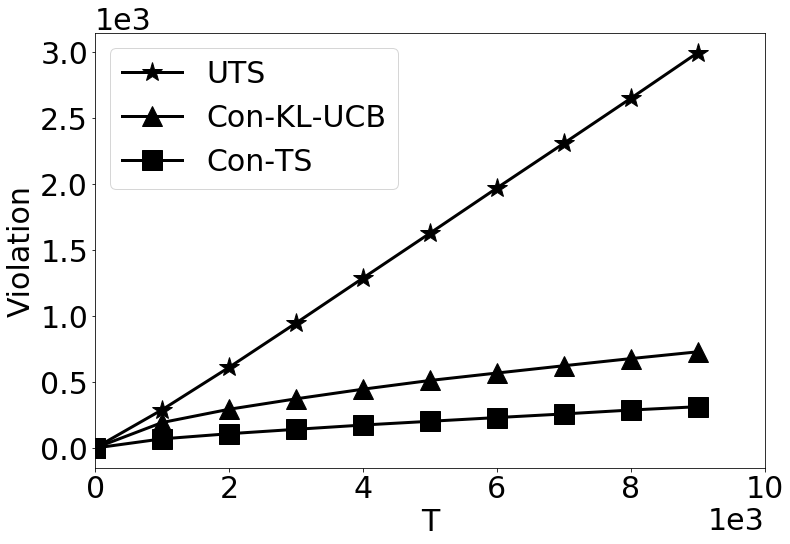}
        \label{fig:vio_linear}
    }
    
\caption{Ratio of cumulative throughput to cumulative violation (top figures, higher is better) and cumulative violation (bottom figures, lower is better) for unimodal Thompson sampling (UTS), constrained KL-UCB (Con-KL-UCB) and the constrained TS (Con-TS).}
\label{fig:WiFi}
\end{figure*}

We evaluate the performance of Con-TS in commonly studied synthetic WiFi scenarios for the standard WiFi data transmission rates. The first three scenarios correspond to packet success probabilities denoted by Gradual, Lossy, and Steep respectively~\cite{combes2014optimal}. The fourth scenario, Linear, uses linearly decreasing packet success probabilities. In each transmission time interval, the packet success events are obtained by sampling a Bernoulli distribution with a mean value equal to the packet success probability for the selected rate. From the field analysis of WiFi hop latencies documented in~\cite{pei2016wifi}, we choose a target packet success probability $\tau=0.75$, for which the average and $90$ percentile end-to-end latencies are reported to be $20$ms and $50$ms respectively. We average the results over $64$ independent experimental runs. The complete simulation parameters are listed in Table~\ref{tab:params}. In addition to the results presented here, numerical results with non-stationary packet success probabilities were obtained by slowly interpolating the packet success probabilities across the Gradual, Lossy and Steep environments. For the non-stationary environment, Con-TS significantly outperforms the UTS as well as the Con-KL-UCB algorithms. The plots and discussion of these results are provided in the Sec.~\ref{app:nonst} owing to space constraints.

All the experiments were carried on a commercial workstation with a $12$-core Intel Xeon processor running at 3.50 GHz and equipped with a single Tesla P100 graphical processing unit. The code for the simulations is written in Python and uses the Numpy and Matplotlib Python libraries for the mathematical operations and the plotting functions respectively. Additionally, we use the Python interface to the CVXOPT library for solving the linear program in each round of the algorithm~\cite{vandenberghe2010cvxopt}. Execution of the code was done through remote execution of a Jupyter notebook~\cite{kluyver2016jupyter}. Multiple, parallel runs of the simulation were distributed across the multiple cores using the Ray package~\cite{moritz2018ray}.

\begin{table}
\caption{Simulation Parameters}
\label{tab:params}
\centering
\begin{tabular}[t]{p{1.5cm}p{6cm}}
\toprule
Parameter & Value\\
\midrule
$K$ & 8 \\
$T$ & 10000 \\
$\tau$ & 0.75 \\
Rates                     & $[6, 9, 12, 18, 24, 36, 48, 54]$ Mbps\\
\multicolumn{2}{l}{Packet Success Probabilities:} \\
Gradual               & $[0.95, 0.90, 0.80, 0.65, 0.45, 0.25, 0.15, 0.10]$\\
Lossy              & $[0.90, 0.80, 0.70, 0.55, 0.45, 0.35, 0.20, 0.10]$\\
Steep & $[0.99, 0.98, 0.96, 0.93, 0.90, 0.10, 0.06, 0.04]$\\
Linear & $[1.00, 0.87, 0.75, 0.62, 0.50, 0.37, 0.25, 0.12]$\\
\bottomrule
\label{tab:par}
\end{tabular}
\end{table}

As discussed in preceding sections, the latency requirement imposes a soft constraint on the packet success rates. Consequently, the important metric to study is the balance between the achieved cumulative throughput and the constraint violations. Consequently, we plot the throughput-violation ratio and the cumulative violation metrics for the studied algorithms. We compare Con-TS with the following algorithms proposed in the literature:

(i) Unimodal Thompson Sampling \emph{UTS}, proposed in~\cite{paladino2017unimodal}, which exploits the unimodal property of throughput as function of rates. UTS is  an extension of the upper confidence bound based G-ORS algorithm for rate selection proposed in~\cite{combes2014optimal}. UTS is an unconstrained algorithm, i.e., it does not admit latency constraints and only optimizes for the maximum expected throughput.

(ii) A constrained upper confidence bound algorithm, \emph{Con-KL-UCB}, which is a Kullback-Leibler (KL) divergence based extension of the ConUCB algorithm proposed in~\cite{chen2018beyond}. ConUCB is a multi-play MAB algorithm proposed in for online weblink selection under a probabilistic constraint. The constraint handled by the original ConUCB algorithm models a minimum acceptable click-through-rate. Since rate selection is a non-combinatorial problem with Bernoulli feedback, using the KL-based upper confidence bounds achieves optimal regret for Bernoulli bandits, and empirically better performance than the index-based ConUCB algorithm~\cite{garivier2011kl}.

In Fig.~\ref{fig:rew_gradual}, Fig.~\ref{fig:rew_lossy}, Fig.~\ref{fig:rew_steep}, and Fig.~\ref{fig:rew_linear}, we plot the ratio of cumulative throughput to cumulative violation for Gradual, Lossy, Steep, and Linear WiFi environments respectively. For all environments except the Steep environment, the proposed Con-TS algorithm achieves a ratio that is approximate double of the next best algorithm for all scenarios. In these scenarios, Con-KL-UCB is the next best algorithm followed by UTS. The performance for each of the algorithms is similar between the Lossy and Linear scenarios, which can be explained by the apparent closeness in the  packet success probabilities for the two environments. For the Gradual scenario, UTS slightly outperforms Con-KL-UCB for the initial few time intervals, but falls behind starting at approximately $T=4000$. For the Steep scenario, the relative performance of the schemes is quite different: UTS performs the best, achieving nearly twice the throughput-violation ratio obtained for Con-TS. The reason for this is apparent on carefully investigating the parameters for Steep scenario. The expected throughput for Steep scenario is maximized by a data transmission rate of $24$ Mbps, which has a true packet success probability of $0.90$. Since the target packet success probability is only $0.75$, the constraint in \eqref{eq:opt_rate_sel_linprog} is trivially satisfied and the problem reduces it to an unconstrained one. For this problem, UTS can better exploit the structure of the rates and consequently achieves a better performance.

In Fig.~\ref{fig:vio_gradual}, Fig.~\ref{fig:vio_lossy}, Fig.~\ref{fig:vio_steep}, and Fig.~\ref{fig:vio_linear}, we plot the  cumulative violations for Gradual, Lossy, Steep, and Linear WiFi environments respectively. For each of these environments, we observe that Con-TS accumulates very few constraint violations. The largest advantage of Con-TS is observed for the Gradual  environment, where Con-TS has less than half cumulative violations compared to the competing algorithms. In contrast, UTS suffers from linear violation rate for the Gradual, Lossy, and Linear scenarios, on account of being an unconstrained algorithm. In case of the Steep scenario, where the problem reduces to an unconstrained version, Con-TS, UTS, and Con-KL-UCB all manage to achieve a relatively small number of constraint violations. Further, these schemes quickly converge to the optimal rate selection policy. 

\section{Conclusions and Future Work}
In this paper, we have proposed Con-TS, a novel constrained TS algorithm for latency-sensitive rate selection in wireless systems. Our algorithms is shown to be upper bounded by $O(\sqrt{T})$ in terms of the cumulative number of violations and the cumulative loss in throughput. We have provided numerical results to compare Con-TS with recently proposed MAB algorithms, both using the TS heuristic as well as the UCB technique. Through these numerical results, we have demonstrated that Con-TS outperforms the state-of-the-art MAB algorithms for several established stationary wireless propagation environments, and an example non-stationary wireless environment.

Additionally in this paper, we have formulated rate selection as a knapsack problem and developed a linear programming relaxation for this problem. We believe that this novel formulation of the rate selection problem opens it up for further theoretical analysis, and allows approaches from other, unrelated, domains to be applied here. Moreover, other important problems in wireless communications that involve probabilistic constraints can benefit from the approach proposed in this paper. One example relates to the selection the optimal beam from a set of candidate transmission beams~\cite{hashemi2018efficient} for multi-antenna transmission, where there are often constraints in terms of the acceptable level of average transmit power.

Finally, we note that rate selection and similar problems often exhibit structure, such as the monotonicity and unimodality properties mentioned in this paper. Further, practical wireless networks often have access to additional side information related to the wireless channel quality. These structural properties as well as the availability of side information may be exploited using suitable TS priors and tailored extensions to the Con-TS algorithms proposed here.

\section*{Acknowledgements}

This work was partially supported by the Wallenberg AI, Autonomous Systems and Software Program (WASP) funded by the Knut and Alice Wallenberg Foundation.

\bibliography{refs}
\bibliographystyle{icml2020}

\newpage

\section{Appendices}
\subsection{Proof of violation bound}
\label{app:proof_vio}

\begin{proof}
Since $p_k(t)$ is the optimal solution to the LP in the $t^\text{th}$ round of Con-TS, it holds almost surely that
\begin{align*}
    \sum_{k=1}^Kp_k(t) \widetilde{\mu}_k(t) \geq \tau.
\end{align*}

    We would like to bound the expected violation
    \begin{align}
    \mathbb{E}\bigg[\bigg[T\tau - & \sum_{t=1}^T\sum_{k=1}^Kp_k(t)\mu_k\bigg]_+\bigg] \nonumber \\
    &\leq \mathbb{E}\bigg[\bigg[\sum_{t=1}^T\sum_{k=1}^Kp_k(t)\widetilde{\mu}_k(t) - \sum_{t=1}^T\sum_{k=1}^Kp_k(t)\mu_k\bigg]_+\bigg] \nonumber \\
    &= \mathbb{E}\bigg[\bigg[\sum_{t=1}^T\sum_{k=1}^K(\widetilde{\mu}_k(t) - \mu_k)p_k(t)\bigg]_+\bigg] \nonumber \\
    &\leq \sum_{t=1}^T\sum_{k=1}^K\mathbb{E}[[\widetilde{\mu}_k(t) - \mu_k]_+].
    \label{eq:violation_bound}
    \end{align}
where we have used the fact that $(a+b)_+\leq a_+ + b_+$ and $p_k(t)\leq 1$. We use the decomposition proposed in~\cite{russo2014learning} in terms of the upper and lower confidence bounds,
\begin{align}
    &\sum_{t=1}^T\sum_{k=1}^K\mathbb{E}[(\widetilde{\mu}_k(t) - U_k(t) + U_k(t)-L_k(t)+L_k(t)-\mu_k)_+] \nonumber \\
    &\leq \sum_{t=1}^T\sum_{k=1}^K\mathbb{E}[(\widetilde{\mu}_k(t) - U_k(t))_+] + \sum_{t=1}^T\sum_{k=1}^K\mathbb{E}[(L_k(t)-\mu_k)_+] \nonumber \\
    &\quad+ \sum_{t=1}^T\sum_{k=1}^K\mathbb{E}[(U_k(t) - L_k(t))].
    \label{eq:vio_deriv}
\end{align}
For the first term, we observed that since $U_k(t)$ is a deterministic function of $t$, we can rewrite the expectation in terms of its conditional expectation on the history $\mathcal{H}(t)$, which leads to
\begin{align}
    \mathbb{E}[\widetilde{\mu}_k(t) - U_k(t))_+] &= \mathbb{E}[\mathbb{E}[(\widetilde{\mu}_k(t) - U_k(t))_+|\mathcal{H}(t)]] \nonumber\\
    &= \mathbb{E}[\mathbb{E}[(\mu_k(t) - U_k(t))_+|\mathcal{H}(t)]]\nonumber\\
    &=\mathbb{E}[\mu_k(t) - U_k(t))_+]\nonumber\\
    &\leq 6\frac{1}{KT}.
\end{align}
The second term in \eqref{eq:vio_deriv} can be bounded directly using Lemma 1, $\mathbb{E}[(L_k(t)-\mu_k)_+]\leq 6\frac{1}{KT}$. Finally, the last term in \eqref{eq:vio_deriv} follows from~\cite{ferreira2018online}[EC.27], $\mathbb{E}[U_k(t) - L_k(t)]=O(K^2\log T\sqrt{T})$. This completes the proof of the violation bound.

\end{proof}

\begin{figure*}[t]
\centering

    \subfigure[M.A. tput  / M.A. vio., nonstationary]{
    \includegraphics[width=0.485\textwidth]{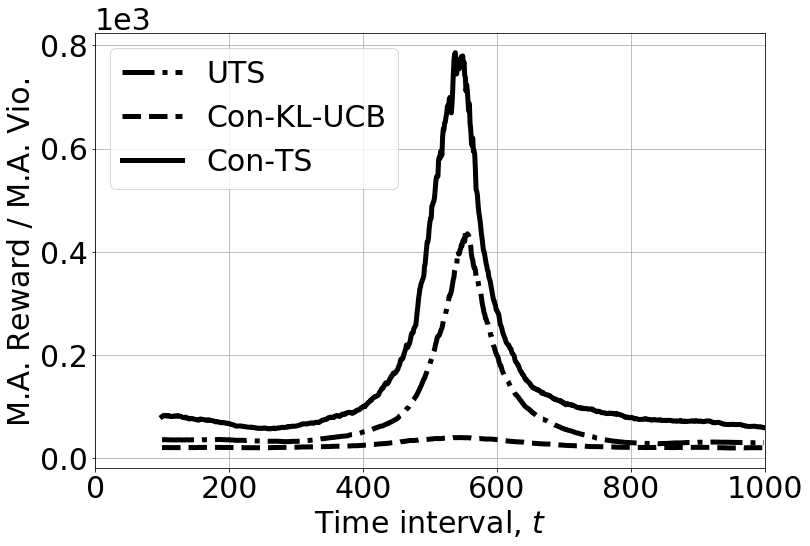}
        \label{fig:rew_nonst_wifi}
    }
    \subfigure[M.A. vio., nonstationary]{
    \includegraphics[width=0.485\textwidth]{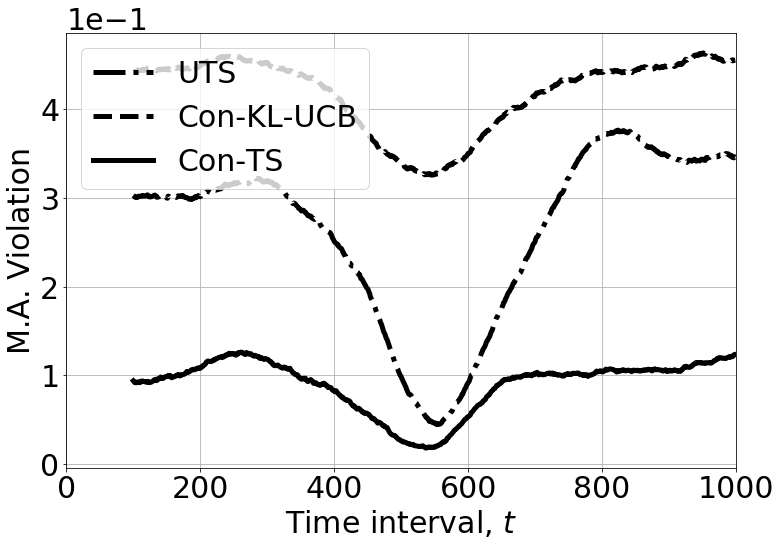}
        \label{fig:vio_nonst_wifi}
    }

\caption{Ratio of cumulative throughput to cumulative violation (higher is better), and cumulation violation (lower is better) for unimodal Thompson sampling (UTS), constrained KL-UCB (Con-KL-UCB) and the proposed Con-TS algorithms.}
\label{fig:WiFi}
\end{figure*}

\subsection{Proof of Regret Bound}
\label{app:proof_regret}

\begin{proof}
We would like to bound the expected regret
\begin{align}
\mathbb{E}\bigg[\bigg[ T\sum_{k=1}^K p_k^*r_k\mu_k - \sum_{t=1}^T \sum_{k=1}^Kp_k(t)r_k\mu_k\bigg]_+\bigg],
\end{align}
We can use the law of iterated expectations to obtain
\begin{align}
    &\mathbb{E}\bigg[\bigg[\sum_{t=1}^T\sum_{k=1}^Kr_k\mu_k(p_k^*-p_k(t))\bigg]_+\bigg] \nonumber \\
    &\qquad= \sum_{t=1}^T\sum_{k=1}^K\mathbb{E}[\mathbb{E}[(r_k\mu_kp_k^*-r_k\mu_kp_k(t))_+|\mathcal{H}_{t}]].
    \label{eq:exp_regret_step_1}
\end{align} 
Similar to the decomposition used in~\cite{russo2014learning}, we can add and subtract the upper confidence bound to get
\begin{align}
    &\sum_{t=1}^T\sum_{k=1}^K\mathbb{E}[\mathbb{E}[r_k(\mu_kp_k^*-U_k(t)p_k^*+U_k(t)p_k(t)-\mu_kp_k(t))_+|\mathcal{H}_{t}]] \nonumber \\
    &\qquad= \sum_{t=1}^T\sum_{k=1}^K\mathbb{E}[\mathbb{E}[r_kp_k^*(\mu_k-U_k(t))_+|\mathcal{H}_{t}]] \nonumber \\
    &\quad\qquad + \sum_{t=1}^T\sum_{k=1}^K\mathbb{E}[\mathbb{E}[r_kp_k(t)(U_k(t)-\mu_k)_+|\mathcal{H}_{t}]].
    \label{eq:exp_regret_step_2}
\end{align}
Now we can reduce the above expression to the sum of two distinct terms, which contain $p_k^*$ and $p_k(t)$ respectively. We then apply the argument from before that $p_k^*$ and $p_k(t)$ are distributed identically conditioned on the history to obtain
\begin{align}
    (\ref{eq:exp_regret_step_2})&=\sum_{t=1}^T\sum_{k=1}^K\mathbb{E}[r_kp_k^*(\mu_k-U_k(t))_+] \nonumber \\
    &\quad + \sum_{t=1}^T\sum_{k=1}^K\mathbb{E}[r_kp_k(t)(U_k(t)-\mu_k)]\\
    &\leq \sum_{t=1}^T\sum_{k=1}^K\mathbb{E}[r_k(\mu_k-U_k(t))_+] \nonumber \\
    &\quad + \sum_{t=1}^T\sum_{k=1}^Kr_k\mathbb{E}[U_k(t)p_k(t)-\widetilde{\mu}_k(t)p_k(t))] \nonumber \\
    &\quad = r_{\max}(6\sqrt{KT} + 12\sqrt{KT\log K}).
    \label{eq:exp_regret_step_3}
\end{align}
where $r_{\max}=\max(\{r_1,\dots,r_K\})$ is the largest available data transmission rate and we have applied Lemma 1 and Lemma 2 to bound the expectations in the penultimate step above.
\end{proof}

\subsection{Numerical Results with Non-stationary Environment}
\label{app:nonst}

The non-stationary environments are generated by linearly interpolating between Gradual, Lossy, and Steep environments over $250$ time intervals, for a total of $T=1000$ time intervals. For non-stationary environments, we use sliding windows spanning $100$ time intervals for each algorithm, and calculate the moving averages (M.A.) of the performance metrics over every $100$ time intervals. The results are provided in Fig.~\ref{fig:WiFi}. Here, Con-TS is observed to outperform both UTS and Con-KL-UCB in terms of the throughput-violation ratio, and the cumulative number of constraint violations.

\end{document}